\documentclass[final]{cvpr}

\usepackage{times}
\usepackage{epsfig}
\usepackage{graphicx}
\usepackage{amsmath}
\usepackage{amssymb}
\usepackage{algorithm}
\usepackage{algorithmicx}
\usepackage{algpseudocode}
\usepackage{booktabs}
\usepackage{array}
\usepackage{colortbl}
\usepackage{overpic}
\usepackage[table]{xcolor}
\usepackage{multirow}

\usepackage{pifont}
\newcommand{\cmark}{\ding{51}}%
\newcommand{\xmark}{\ding{55}}%

\def\ie{\emph{i.e.,~}}
\def\eg{\emph{e.g.,~}}

\newcommand{\sorb}[1]{{\textcolor[rgb]{0.72,0.00,0.00}{\textbf{#1}}}}

\newcommand{\sobb}[1]{{\textcolor{blue}{\textbf{#1}}}}

\newcommand{\myPara}[1]{\vspace{2mm}\noindent\textbf{#1}}

\definecolor{mygreen}{RGB}{0,150,0}
\definecolor{myred}{RGB}{200,0,0}
\usepackage[pagebackref,breaklinks,colorlinks]{hyperref}

\def\cvprPaperID{1853} 
\def\confYear{CVPR 2021}

\graphicspath{{./figures/}}

\newlength\savedwidth
\newcommand{\whline}[1]{\noalign{\global\savedwidth\arrayrulewidth \global\arrayrulewidth #1}%
                   \hline \noalign{\global\arrayrulewidth\savedwidth}}

\begin{document}
	
\title{Centralized Information Interaction for Salient Object Detection}
	
\author{Jiang-Jiang Liu\thanks{Indicates equal contributions.}
    \quad  \quad Zhi-Ang Liu\footnotemark[1] \quad \quad 
    Ming-Ming Cheng\thanks{Indicates corresponding author.} \\
    CS, Nankai University \\
}

\maketitle
\thispagestyle{empty}

\begin{abstract}
The U-shape structure 
has shown its advantage in salient object detection  
for efficiently combining multi-scale features.
However, most existing U-shape based methods focused on improving 
the bottom-up and top-down pathways while ignoring the connections between them.
This paper shows that by centralizing these connections, 
we can achieve the cross-scale information interaction among them, 
hence obtaining semantically stronger and positionally more precise features.
To inspire the potential of the newly proposed strategy, 
we further design a relative global calibration module 
that can simultaneously process multi-scale inputs
without spatial interpolation.
Benefiting from the above 
strategy and module, 
our proposed approach 
can aggregate features more effectively  
while introducing only a few additional parameters.
Our approach can cooperate with various existing 
U-shape-based salient object detection methods 
by substituting the connections between 
the bottom-up and top-down pathways.
Experimental results demonstrate that our proposed approach
performs favorably against the previous state-of-the-arts 
on five widely used benchmarks with less computational complexity.
The source code will be publicly available.

\end{abstract}

\section{Introduction} \label{sec:introduction}

As a fundamental component of low-level computer vision 
and benefiting from its category-agnostic character,
salient object detection has been widely applied in various downstream vision tasks,
such as 
weakly supervised semantic segmentation \cite{wei2016stc,hou2018self}, 
visual tracking \cite{hong2015online},
content-aware image editing \cite{cheng2010repfinder},
and robot navigation \cite{craye2016environment}.
%
%
Traditional salient object detection methods depend heavily on hand-crafted feature detectors.
%
%
These detectors
cannot utilize the rich high-level semantic information hidden in the 
image and dataset, making them fail in complex scenes.
%
%
%
With the popularization of commercial GPUs,
convolutional neural networks (CNNs) based methods have been developing rapidly
in recent years for their capability of extracting both high-level semantics 
and low-level textures of multiple scales.
%
%
%

\begin{figure}[tp]
    \centering
    \includegraphics[width=0.95\linewidth]{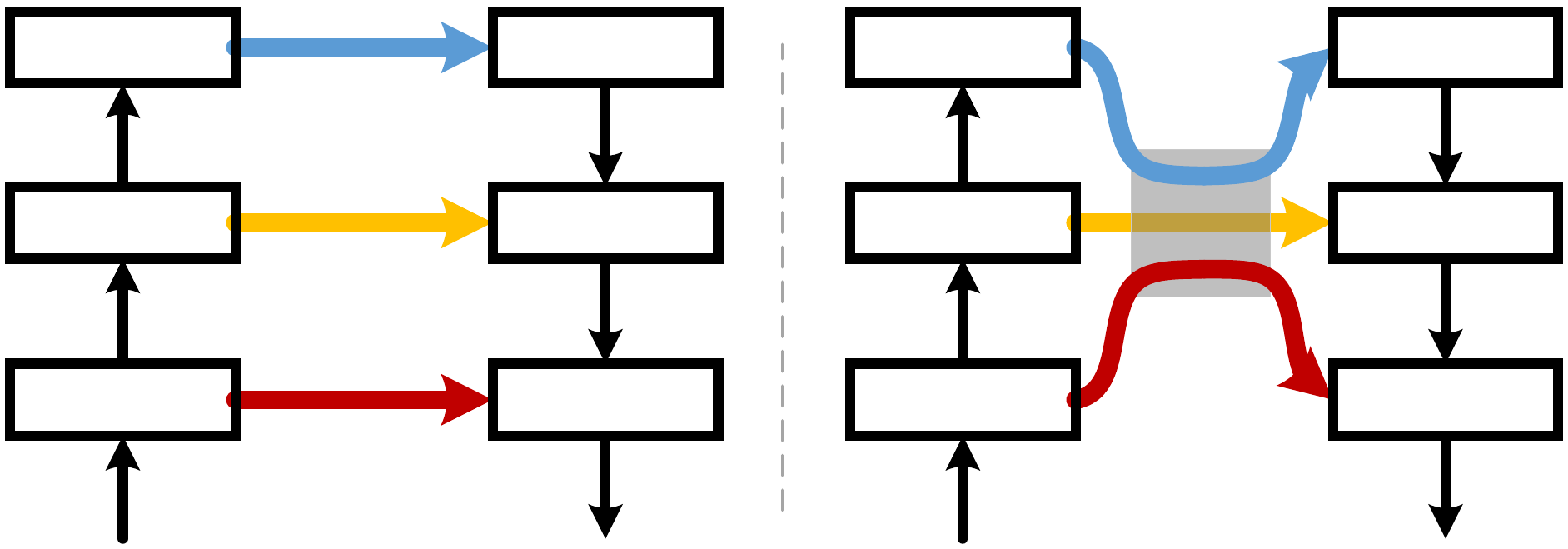}
	\caption{Conceptual diagram of our proposed 
	centralized information interaction strategy.
	\textbf{Left}: Typical U-shape structure 
	which connects the corresponding stages in the bottom-up and 
	top-down pathways directly;
	\textbf{Right}: Our proposed centralized strategy. 
	The rounded gray rectangle refers to a 
	shared module that can parallelly process all stages (scales) of features
	in a stage-wise manner. 
	}\label{fig:teaser}
    \vspace{-4pt}
\end{figure}

One representative architecture for salient object detection 
is the U-shape structure \cite{ronneberger2015u,lin2017feature}.
%
%
As illustrated in the left part of Fig.~\ref{fig:teaser}, 
a typical U-shape structure consists of a bottom-up pathway, 
a top-down pathway, and several connections between them. 
%
Among the methods that aim to advance the U-shape structures,
most of them either focus on 
improving the bottom-up pathway's feature extraction capability or 
enhancing the top-down pathway's feature aggregation ability or both of them.
%
%
However, the connections between the bottom-up and top-down pathways
are neglected.
The usual practice is to directly connect the corresponding stages.
In this paper, different from the methods mentioned above,
we investigate how to augment the representation capability 
of the extracted features    
by redesigning the connections 
between the bottom-up and top-down pathways 
rather than the pathways themselves.

\begin{figure}[tp]
    \centering
    \includegraphics[width=1.\linewidth]{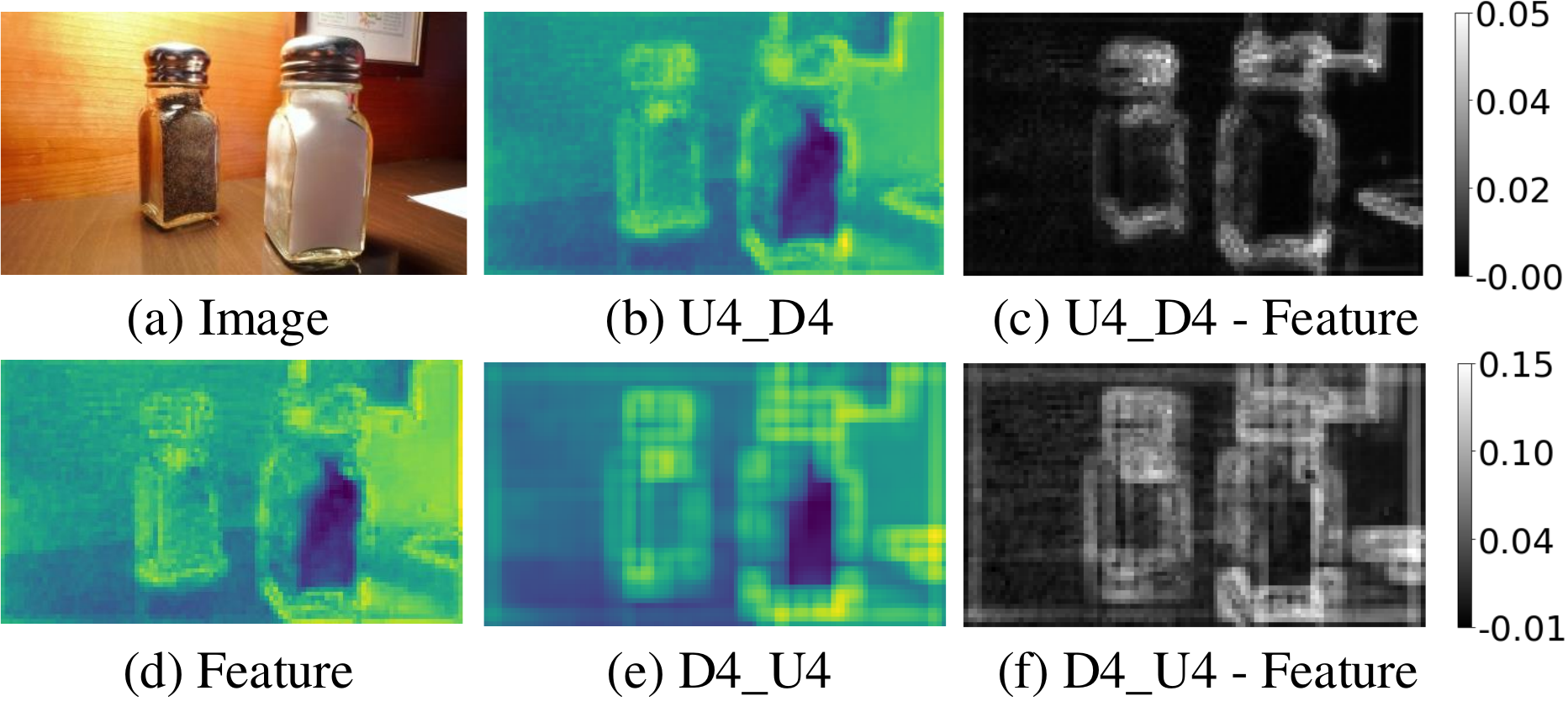}
	\caption{An example case showing the impact of 
	spatial interpolation. (a) is the source image and 
	(d) is the corresponding output feature map 
	of ResNet-18's \texttt{res1} stage. 
	(b) is obtained by directly $4 \times$ bilinear up-sampling and then 
	$4 \times$ bilinear down-sampling on (d), while (e) is down-sampled
	first and then up-sampled.
	(c,f) are the difference maps of (b,d) and (e,d), respectively. 
	It can be seen that even the simplest spatial interpolation operations 
	can cause obvious differences.
    }\label{fig:interpolation}
\end{figure}

A straightforward way to achieve the 
purpose mentioned above is to
fuse the extracted multi-scale features directly
\cite{pang2019libra,li2020cross}. 
However, an inevitable step of cross-scale features fusing 
is spatial interpolation. 
As demonstrated in Fig.~\ref{fig:interpolation}, 
the intermediate feature map being down-sampled 
first and then up-sampled differs greatly from its original values,  
and vice versa.
This phenomenon worsens when the down-sampling rate increases, 
as more spatial location information is missed.

To this end, we propose to encode the cross-scale information
into the filters instead of the features, where no spatial interpolation is required. 
%
As shown in the right part of Fig.~\ref{fig:teaser}, 
the multi-scale features extracted from the bottom-up pathway are 
parallelly processed by a centralized module whose parameters are shared 
across scales before being
used to build the top-down pathway.
%
%
We call the proposed filter-level information interaction strategy,
centralized information interaction (CII).
CII faithfully preserves the exact spatial locations within each input scale, 
meanwhile achieves cross-scale information interaction 
and fundamentally avoids the negative effects caused by spatial  interpolation of features.
%
%
%
%
To cooperate with the new normal that 
the inputs of the shared module in CII are naturally of multiple scales,    
we further propose a relative global calibration module (RGC).
By exploring the global information 
related to each scale of feature,
RGC achieves a balance between 
essential global semantic and local details. 
%
We also show that with a slight modification 
on the input flow of the RGC module, 
the overall performance can be further promoted nearly for free.

To evaluate the performance of the proposed approach, 
we report results on five popular salient object detection benchmarks.
%
We conduct extensive ablation studies and 
show numerous visual examples to help 
readers better understand the impacts
of different components of the proposed approach.
%
%
Our approach only introduces $\sim$ 4\% additional parameters 
regarding the backbone network.
It can be trained end-to-end on a single RTX-2080Ti GPU within 1.5 hours
on a set of 10,553 images.
%
%
%
To sum up, our main contributions can be summarized as follows:
\begin{itemize}
	\item We design a centralized information interaction strategy 
	to efficiently resolve the contradiction between 
	cross-scale information interaction 
	and spatial location information preservation. 
	\item We propose a relative global calibration module that can effectively
	exploit the relative global information related to each different spatial scale
	and obtain substantial improvements.
\end{itemize}

\section{Related Work}

In this section, we briefly review the recent representative work 
on salient object detection, 
multi-scale feature aggregation, and multi-scale and attention modules. 

\subsection{Salient Object Detection} 
Early salient object detection methods were usually 
based on intrinsic cues and 
hand-crafted features\cite{perazzi2012saliency,jiang2013salient,cheng2015global,li2013saliency}. 
More details can be found in 
recent surveys \cite{borji2015salient,BorjiCVM2019,wang2019revisiting,wang2019survey}. 
%
%
%
%
%
Among the deep-learning-based methods, 
many adopted the idea of recurrent refinement 
\cite{wang2018detect,wang2019iterative}
to refine the predictions iteratively.
%
Some methods treated this problem stage-wisely  
\cite{liu2016dhsnet,wang2017stagewise,wu2019cascaded}
by decoupling it into multiple stages. 
%
To get predictions with more precise boundaries, 
\cite{qin2019basnet,zhou2020interactive,F3Net} 
designed new loss functions 
while \cite{feng2019attentive,zhao2019egnet,zhou2020interactive,wei2020label} introduced extra supervisions.
%
\cite{li2017instance,liu2018picanet,zhang2018progressive,wang2019salient,feng2019attentive,zhao2019pyramid}
proposed various types of attention mechanisms 
and achieved substantial improvements.
%
%
Among the above methods, a majority of them were 
based on the classic U-shape structure.
\cite{zhang2018bi,zhao2019pyramid,su2019selectivity,zeng2019towards, GateNet2020} 
attached additional multi-scale modules 
after the bottom-up pathway  
to generate more powerful features,
while \cite{hou2016deeply,zhang2017amulet,luo2017non,liu2019simple,pang2020multi} combined the extracted multi-scale features 
in different ways within the top-down pathway 
to generate richer features.
\begin{figure*}[t!]
    \centering
    \includegraphics[width=1.0\textwidth]{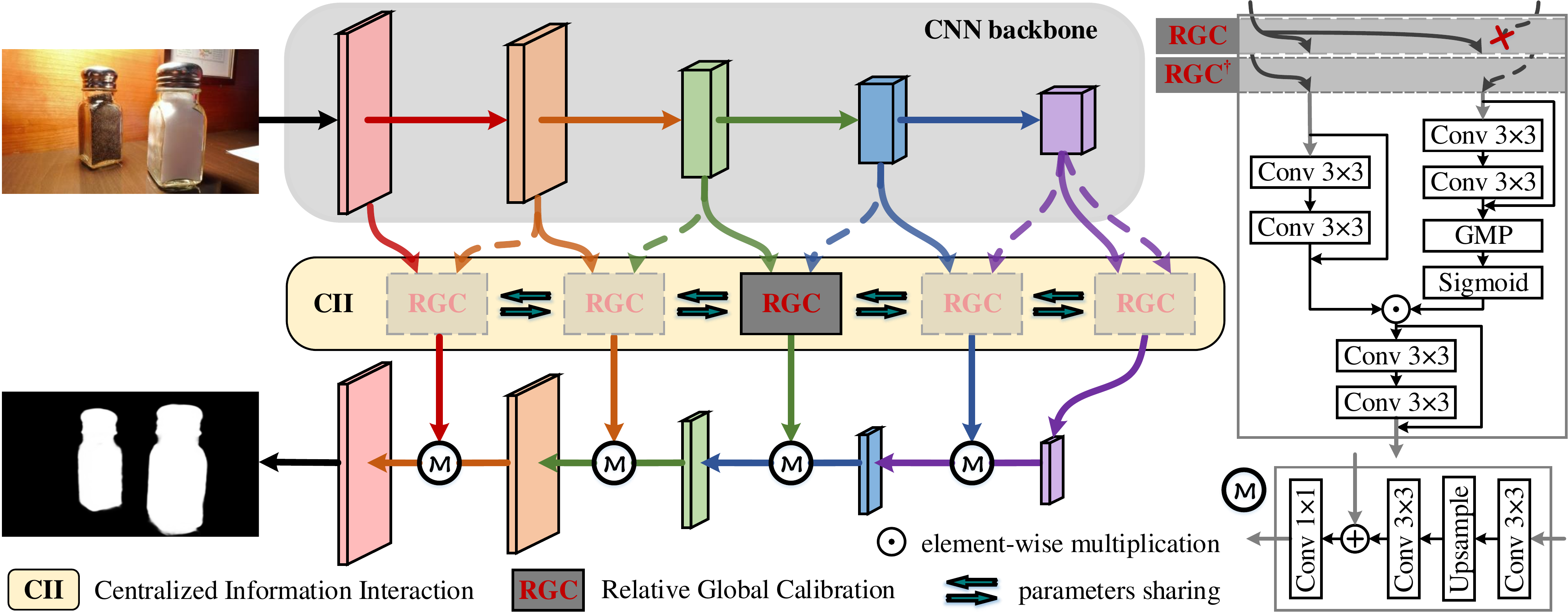}
	\caption{The overall pipeline of our proposed approach. 
	Note that 
	the dashed arrows only active in $\textbf{RGC}^\text{\dag}$, 
	as shown in the top-right corner. }
	\label{fig:pipeline}
	\vspace{-4pt}
\end{figure*}

\subsection{Multi-scale Features Aggregation} 
%
%
As a pioneer, FCNs \cite{long2015fully} 
directly aggregated the features from 
low-level stages with the ones from the most
high-level stage for more precise semantic segmentation. 
Similarly, U-Net \cite{ronneberger2015u} and FPNs \cite{lin2017feature} 
further incorporated a top-down pathway to sequentially combine the extracted
multi-scale features from high- to low-levels.
%
As followers, PANet \cite{liu2018path} adopted another bottom-up pathway 
on top of FPN while ASFF \cite{liu2019learning} proposed to fuse more stages 
of features in the top-down pathway of FPN.
EfficientDet \cite{tan2020efficientdet} proposed a BiFPN layer and 
repeated it multiple times. 
RFP \cite{qiao2020detectors} proposed to repeatedly pass the features
through the bottom-up backbone to enrich
the representation power of FPN.
Recently, NAS-FPN \cite{ghiasi2019fpn} and Auto-FPN \cite{xu2019auto} applied the 
neural architecture search \cite{zoph2016neural} to discover 
the optimal FPN structure in a data-driven manner automatically.
%


\subsection{Multi-scale and Attention Modules}
%
Deeplabv2 \cite{chen2017deeplab} proposed an 
atrous spatial pyramid pooling (ASPP) module to 
capture contextual information 
using different dilation convolutions.
DenseASPP \cite{yang2018denseaspp} improved  
ASPP with dense connections.
PSPNet \cite{zhao2016pyramid} utilized a pyramid 
pooling module (PPM) to aggregate contextual information
of multiple scales with pooling operations.
Recently, Auto-Deeplab \cite{liu2019auto} proved that 
optimal multi-scale modules could be 
automatically obtained with 
neural architecture search.
%
%
CBAM \cite{woo2018cbam} proposed a sequence of channel attention 
and spatial attention to augment the input features. 
OCNet \cite{yuan2018ocnet} adopted self-attention mechanism 
to augment ASPP with stronger context extraction capability.
DANet \cite{fu2019dual} utilized
a parallelism of position attention and channel attention 
to model long-range dependencies. 
CCNet \cite{huang2020ccnet} developed 
a criss-cross attention module to 
capture the global context in both 
horizontal and vertical directions.
%

Different from the above-mentioned modules that use 
single-scale input, our proposed RGC module takes 
multi-scale inputs simultaneously.
%
Also, our proposed CII aims to augment the 
information interaction among the connections between the 
bottom-up and top-down pathways of the U-shape structure 
rather than the pathways themselves.

\section{Method}
%
%
In this section, we first  
introduce the overall pipeline of the proposed 
approach. 
We then detailedly describe the two main components 
of the proposed approach, including an information interaction
strategy and a feature calibration module.


\subsection{Overall Pipeline} \label{sec:pipeline}
The proposed approach is based on the widely used U-shape structure, 
which consists of a bottom-up pathway for multi-level feature extraction 
and a top-down pathway to combine them.
%
%
%
%
As illustrated in Fig. \ref{fig:pipeline}, 
%
the multi-scale features extracted from the bottom-up pathway
are parallelly forwarded through the 
information interactors (solid gray rectangles) stage-wisely. 
%
%
We share the parameters of these information interactors 
to achieve efficient cross-scale information interaction by 
learning more powerful filters.
%
Then the interacted features are progressively used to build 
the top-down pathway from high- to low-levels.
%
We call the above information interaction strategy that 
encodes the multi-scale information into the shared filters 
as centralized information interaction (CII).
%
%
Considering that the input features of CII are now of multiple scales, 
we further introduce a relative global calibration (RGC) module to 
cooperate with it.
%
%
RGC achieves a balance between essential global semantics 
and local textures by adaptively exploiting 
the relative global information concerning each different input scale.
%
In the following subsections, 
we describe the above-mentioned strategy and module in detail.

\subsection{Centralized Information Interaction} \label{sec:strategy}
One typical design in the classic U-shape structure is the 
short connections between the stages of the same spatial scales 
from the bottom-up pathway to the top-down pathway.
This design provides a simple and efficient way to 
combine the extracted multi-level feature maps. 
Taking the ResNet-18 \cite{He2016} version of the classic U-shape structure for example,
the feature maps outputted by \texttt{conv1, res1, res2, res3, res4} which are
denoted by $\mathbb{B} = \{B_i\}$ ($1\le i \le M$ and $M=5$) are usually used to build 
the output feature pyramid of 
the bottom-up pathway.
%
%
In the top-down pathway, 
the most high-level feature map 
is progressively up-sampled and then 
aggregated with the feature map of 
the corresponding down-sampling rate. 
As can be noticed, 
feature map $B_i$ can not get information from feature maps 
belonging to higher stages $B_j$ ($ 1 \le i < j \le M$) until 
it is aggregated with $B_{i+1}$ in the top-down pathway.
Suppose we treat the
connections between the bottom-up and top-down pathways 
as an independent part. 
In that case, 
the information flows of different scales inside this part are 
independent and unknown to each other. 
%

%
The insightful point that 
lower-level features contain more local textures and patterns
while the higher-level ones indicate the locations of the 
entire target objects manifests the necessity of 
augmenting the interaction among them to 
complement all features with 
more accurate localization and more precise segmentation capabilities.
To this end, we propose to enhance the 
information interaction among the multi-scale feature maps 
when they are 
transmitted from the 
bottom-up to the top-down pathways.
%

As shown in the rounded yellow rectangle of Fig.~\ref{fig:pipeline},  
compared to the classic U-shape structure that directly delivers 
the extracted multi-scale 
feature maps to the top-down pathway,
CII utilizes a series of 
identical information interactors (solid gray rectangles) 
to interact the information encoded in them. 
These information interactors are placed 
at the center of the classic U-shape structure and 
the parameters of them are shared.
In this way, by being encoded into the shared learnable filters, 
the multi-scale informations can interact with each other.
%
%
%
Note that CII is not a specific module but a strategy.
The design of the information interactors in CII is flexible 
and can be replaced with 
various successful modules 
\cite{zhao2016pyramid,chen2017deeplab,yang2018denseaspp,woo2018cbam} 
to have inputs and outputs of identical shapes.
%

%
When basing on the ResNet-18 backbone, 
the channel numbers corresponding to $\mathbb{B}$ 
are set to $\{64, 64, 128, 256, 512\}$, respectively.
We apply a $1 \times 1$ convolution layer ($f_i^{1 \times 1}$) 
after each $B_i \in \mathbb{B}$ to map the input channels 
to the same output channel (\ie 64).
Note that in the following parts of the paper,
we omit the batch normalization layer (BN \cite{ioffe2015batch}) and 
non-linear activation layer (ReLU \cite{nair2010rectified}) after each convolution layer
for notational convenience.
The mapped feature maps are 
processed by the information interactors to 
obtain the output feature maps of CII: 
$\mathbb{C} = \{C_i\}$ ($1\le i \le M$ and $M=5$), respectively.
The overall process of CII can be summarized as:
\begin{equation}
	C_i = InI_i(f_i^{1 \times 1} (B_i)), ~1 \le i \le M, \label{equ:cii}
\end{equation}
where $C_i$ and $B_i$ are of the same spatial shapes, 
and $InI_i$ refers to the identical information interactors whose 
parameters are shared for every $i$.
$\mathbb{C}$ are then used to build the top-down pathway. 
%

Quite different from the previous methods \cite{pang2019libra,li2020cross}
that achieve information interaction 
by directly fusing the multi-scale features 
(\eg concatenation or summation), 
our CII encodes the information into the shared learnable filters.
%
The information interactors in 
CII can get optimization signals from both high- and low-level features,
resulting in semantically stronger and positionally more precise patterns. 
One advantage of CII is that 
the aliasing effect of up-sampling is fundamentally avoided, 
as the input and output feature maps of each information interactor
are of the same spatial sizes, 
indicating that no spatial interpolation operation is required.
Also, CII only introduces a small number of additional parameters 
since we do not need individual 
modules for each scale of inputs.

\begin{figure}[t!]
    \centering
    \includegraphics[width=1.\linewidth]{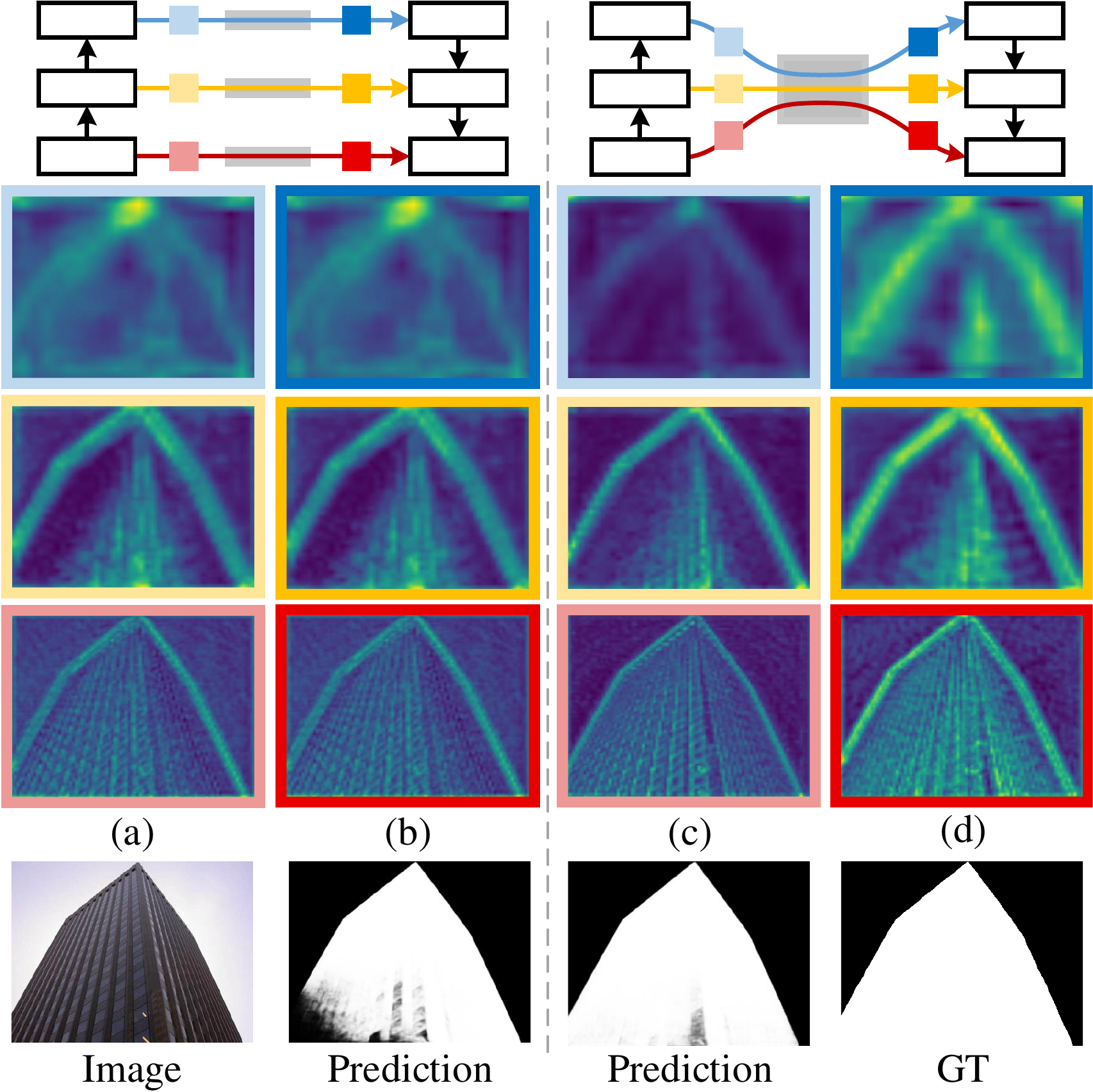}
	\caption{
		Visualizing feature maps w/o and w/ CII. 
		Every translucent gray rectangle refers to a sequence of two $3 \times 3$ 
		convolution layers. 
		As can be seen, CII can interact the 
		complementary information encoded 
		in feature maps of different scales (c,d).
		For comparison, w/o CII, the feature maps are roughly unchanged (a,b).}
	\label{fig:vis_cii}
	\vspace{-4pt}
\end{figure}

To have a straightforward perception,
we show some visual comparisons of 
the intermediate feature maps around CII in Fig.~\ref{fig:vis_cii}.
For simplicity and 
without loss of generality, 
we use a sequence of two $3 \times 3$ convolution layers 
as the information interactors (parameters are shared) in this section 
to build CII.
If we leave the information interactors independent of each other 
(parameters not shared), 
we obtain a simple basic U-shape structure baseline. 
%
By comparing the 4th with the 3rd columns of Fig.~\ref{fig:vis_cii}, 
we can see that after CII, the lower-level features tend to highlight 
more structural pixels than only the exact edge pixels.
The higher-level features, on the contrary, become more detailed 
around the objects' boundaries.
In contrast, without CII, 
the feature maps after and before 
the information interactors are roughly the same
(the 2nd \emph{v.s.} 1st columns in Fig.~\ref{fig:vis_cii}).
This phenomenon verifies the significant effect 
of our CII on complementing the information across scales.
%


		

\subsection{Relative Global Calibration} \label{sec:module}
CII introduces a new strategy for efficient cross-scale 
information interaction against the classic U-shape structure.
As aforementioned, 
when building the information interactors,
a simple sequence of two $3 \times 3$ convolution layers 
outperforms its baseline version with 
a considerable margin (the 3rd \emph{v.s.} 2nd rows in Table.~\ref{tab:cii}).

It is well-known in segmentation-like tasks that 
an effective multi-scale module would always promote the overall performance.
For example, 
the famous PPM \cite{zhao2016pyramid} comprises four 
parallel pooling branches with different down-sampling rates 
to utilize the input feature maps' multi-scale information.
It was first introduced in 
semantic segmentation and has been successfully adopted in many 
salient object detection methods \cite{wang2017stagewise,liu2019simple}.
Based on this prior and that 
PPM is designed to 
be plug-and-play, 
we try to use it as 
the information interactors in CII ($InI_i$ in Equ.~\ref{equ:cii}).
However, it turns out that PPM 
does not work well
with CII (the 4th \emph{v.s.} 1st rows in Table.~\ref{tab:rgc}).
%
%

PPM was designed to collect information from multiple receptive field sizes  
when the input is single-scale.
When the inputs are of multiple scales (\ie $\mathbb{B}$), 
it results in a rapid growth of 
receptive fields sizes (\ie $1 \times 4$ \emph{v.s.} $M \times 4$).
However, as pointed out in a lot of previous literature 
\cite{chen2017deeplab,zhao2016pyramid,liu2020improving},
more diversity does not necessarily mean better results, 
which can even be problematic as 
too much diversity may distract the following layers.
%
%

%
%

%
Considering that the inputs of CII (\ie $\mathbb{B}$) naturally have
multiple receptive field sizes, 
it is essential to cut off the redundancy and 
retain only the necessary diversity.
%
We propose a relative global calibration module that 
contains two parallel branches responsible for 
local information retainment and 
relative global information compaction, 
as illustrated in the top-right part of Fig.~\ref{fig:pipeline}. 
Specifically,  
in both the two branches of RGC, 
$B_i \in \mathbb{B}$ is first processed by 
a sequence of two $3 \times 3$ convolution layers 
(denoted as $f_{L_2}^{3 \times 3}$ and $f_{R_2}^{3 \times 3}$, respectively).
The learnable parameters in these layers 
leave moderate rooms for feature adjustment as
the spatial scales of the feature maps in $\mathbb{B}$ differ.
%
%
Then a global max pooling layer (GMP) is applied after the convolution layers
in the right branch
to compact the relative global information $G_i$ concerning $B_i$: 
\begin{equation} 
	G_i = \sigma (GMP((f_{R_2}^{3 \times 3} +1) (B_i))) , ~1\le i \le M, \label{equ:rgc_gmp}
\end{equation}
where $\sigma$ refers to the sigmoid function.
After that, 
the compacted global information from the right branch is 
used to calibrate the retained local feature from
the left branch.
With another sequence of two $3 \times 3$ convolution layers 
($f_{F_2}^{3 \times 3}$), we can obtain the output $R_i$:
%
%
\begin{equation} 
	R_i = (f_{F_2}^{3 \times 3} + 1) (
		G_i
		\odot (f_{L_2}^{3 \times 3} + 1 ) (B_i)), ~1\le i \le M. \label{equ:rgc_overall}
\end{equation}
Note that all the learnable parameters in Equ:~\ref{equ:rgc_gmp} and \ref{equ:rgc_overall}
are shared for every $i$.

We will show in Sec.~\ref{sec:experiments} that 
when cooperating with CII, 
though with fewer branches, 
RGC outperforms the previous multi-scale modules.
To investigate the potential of RGC, we 
make a small modification to the inputs of RGC 
to lead in global information of larger 
receptive fields without 
interpolating the inputs spatially.
%
By simply replacing the input of the right branch with 
the feature map from its succeeding stage (\ie $B_i$ to $B_i+1$), 
we get $\text{RGC}^\text{\dag}$ 
which can further improve the performances  
while introducing no additional parameters and 
having even less computation complexity.
We will provide more quantitative analysis in the experiment section.

%
%



\section{Experiments} \label{sec:experiments}

\subsection{Experiment Setup}

\myPara{Datasets: }
For all the experiments, 
the DUTS-TR \cite{wang2017learning} dataset 
is used for training as commonly done. 
For performance evaluation, 
five popular datasets: ECSSD \cite{yan2013hierarchical}, 
PASCAL-S \cite{li2014secrets}, 
DUT-OMRON \cite{yang2013saliency}, 
HKU-IS \cite{li2015visual}  
and DUTS-TE \cite{wang2017learning} are used.

\newcommand{\addFig}[1]{\includegraphics[width=0.0885\linewidth]{comparison/#1}}
\newcommand{\addFigs}[1]{\addFig{#1.jpg} & \addFig{#1_gt.png} & \addFig{#1_ours.png} &
			\addFig{#1_MINet.png} & \addFig{#1_ITSD.png} & \addFig{#1_CSNet.png} & \addFig{#1_PoolNet.png} &
			\addFig{#1_BASNet.png} & \addFig{#1_CPD.png} & \addFig{#1_PiCANet.png} & \addFig{#1_Amulet.png}}
\begin{figure*}
	\centering
    \footnotesize
	\setlength\tabcolsep{0.2mm}
	\renewcommand\arraystretch{0.8}
	\begin{tabular}{ccccccccccc}
		\addFigs{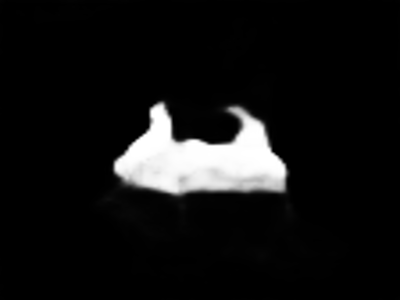}\\
		\addFigs{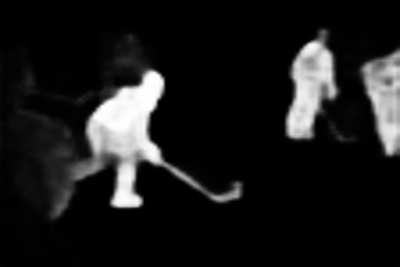}\\
		\addFigs{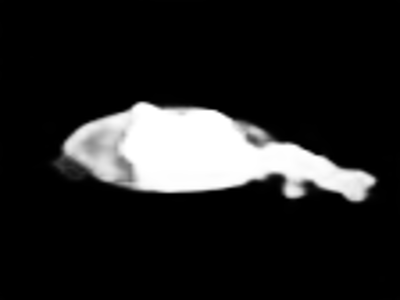}\\
		\addFigs{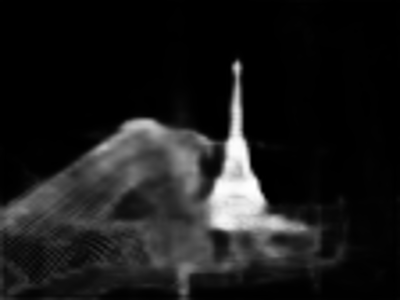}\\
		\addFigs{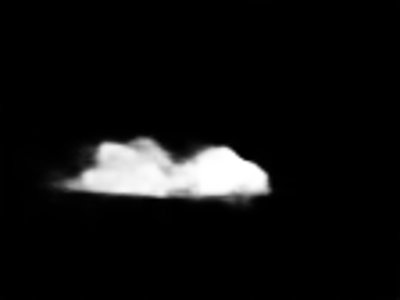}\\
		\addFigs{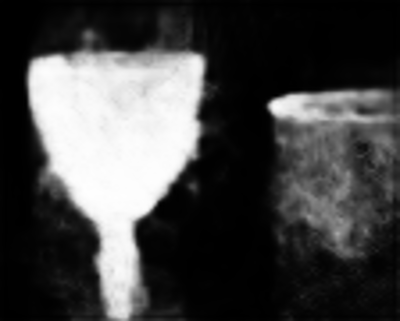}\\
	   Image & GT & Ours & MINet\cite{pang2020multi} & ITSD\cite{zhou2020interactive} & CSNet\cite{gao2020sod100k} &
	   PoolNet\cite{liu2019simple} & BASNet\cite{qin2019basnet} & CPD\cite{wu2019cascaded} &
	   PiCANet\cite{liu2018picanet} & Amulet\cite{zhang2017amulet} \\
	\end{tabular}
  \vspace{-1pt}
	\caption{Qualitative comparisons to previous state-of-the-art methods.
	Compared to other methods, our approach 
	can not only locate the salient objects 
	with cluttered backgrounds but also produce more integral saliency maps.}
	\label{fig:vis_comps}
  \vspace{-8pt}
\end{figure*}

\myPara{Loss Function: }
We utilize the summation of binary cross entropy (BCE) loss 
and intersection over union (IoU) loss as our loss function: 
\begin{equation}
\setlength{\abovedisplayskip}{4pt}
\setlength{\belowdisplayskip}{4pt}
	l = l_{bce} + l_{iou}.
\end{equation}
BCE loss is broadly used in binary classification 
and segmentation tasks due to its robustness, 
which accumulates per-pixel loss in images:
\begin{equation}
\setlength{\abovedisplayskip}{4pt}
\setlength{\belowdisplayskip}{4pt}
	l_{bce}(x,y)=-\frac{1}{n} \sum_{k=1}^n [y_klog(x_k)+(1-y_k)log(1-x_k)], 
\end{equation}
where $x$ and $y$ denote the predicted map 
and the ground truth respectively, 
while $k$ is the index of pixels and $n$ is the number of pixels in $x$.
Different from BCE loss that focuses on the pixel-level differences, 
IoU loss takes into account 
the similarity of the whole image, 
which is defined as follows:
\begin{equation}
\setlength{\abovedisplayskip}{4pt}
\setlength{\belowdisplayskip}{4pt}
	l_{iou}(x,y) = 1-\frac{\sum_{k=1}^n (y_k*x_k)}{\sum_{k=1}^n (y_k+x_k-y_k*x_k)}.
\end{equation}

\myPara{Evaluation Criteria: }
We evaluate the performance of our approach and 
other methods using four widely-used metrics: 
precision-recall (PR) curves, F-measure score ($F_\beta$), 
S-measure score ($S_\alpha$) \cite{fan2017structure},  
and mean absolute error ($MAE$).
The F-measure($F_\beta$) score is formulated as the 
weighted harmonic mean of the average precision and average recall:
\begin{equation}
\setlength{\abovedisplayskip}{8pt}
\setlength{\belowdisplayskip}{8pt}
	F_\beta = \frac{(1+\beta^2) \times Precision \times Recall}{\beta^2 \times Precision + Recall }.
\end{equation}
We set $\beta^2$ to 0.3 to weigh precision more than recall 
as the previous works suggested. 
The S-measure ($S_\alpha$) score reflects both the 
object-aware ($S_o$) and the region-aware ($S_r$) structure similarities 
between the predicted map and the ground truth:
\begin{equation}
	S_\alpha = \gamma S_o + (1-\gamma)S_r,
\end{equation}
where $\gamma$ is set as 0.5 by default.
The $MAE$ score evaluates the average pixel-level relative error
between the normalized predicted map $P$ and ground truth $G$:
\begin{equation}
\setlength{\abovedisplayskip}{4pt}
\setlength{\belowdisplayskip}{4pt}
	MAE = \frac{1}{W \times H} \sum_{x=1}^W \sum_{y=1}^H \lvert P(x,y) - G(x,y) \rvert,
\end{equation}
where $W$ and $H$ denote the width and height of $P$, respectively.

\myPara{Implementation Details: }
We implement our approach using the publicly available PyTorch 
library\footnote{https://pytorch.org} 
and a RTX-2080Ti GPU is used for acceleration. 
The parameters of our backbone network 
(\ie ResNet-18 and ResNet-50 \cite{He2016})
are initialized with the corresponding models pretrained on 
the ImageNet dataset \cite{krizhevsky2012imagenet} 
and the rest are randomly initialized.
For all the experiments, our model is trained for 32 epochs 
with a batch size of 30.
The stochastic gradient descent (SGD) optimizer 
with a momentum of 0.9 and weight decay of 5e-5 
is used for optimization. 
The maximum learning rates are set as 0.005 for the backbone 
and 0.05 for the rest.
We apply warm-up and cosine schedule  
to the learning rates in the first 8 and last 24 epochs, respectively. 
%
%
Random horizontal flipping and random cropping are used for data augmentation. 
In both training and testing, the input images are resized to $352\times352$.
By default, our ablation experiments are 
based on the ResNet-18 \cite{He2016} backbone 
unless special explanations.  
Note that we do not apply any pre- or post-processing techniques. 
%
%

\subsection{Ablation Studies}
In this subsection, 
we first conduct several straightforward experiments to 
show the effectiveness of the purposed CII and RGC 
from an overall perspective.
Then we detailedly investigate the design choices 
and analyze the configurations 
of both CII and RGC with more ablation studies. 
%
%

\myPara{Effectiveness of CII:} 
To demonstrate the effectiveness of CII  
against the classic U-shape structure, 
we compare different settings of the connections between 
the bottom-up and top-down pathways in Table.~\ref{tab:cii}.
Except for the settings of the connections, 
all other configurations are identical.
%
The 1st row in Table.~\ref{tab:cii} is the U-shape baseline which connects 
the bottom-up and top-down pathways with a 
single $1 \times 1$ convolution layer, respectively.
%
As a comparison, 
the 3rd row applies our proposed CII with 
a sequence of two $3 \times 3$ convolution layers 
as the information interactor ($InI_i$ in Equ.~\ref{equ:cii}), 
which is shared across scales.
As can be seen,  
the utilization of CII greatly promotes the overall performances 
compared to the U-shape baseline.
%
\begin{itemize}
	\item
	\textbf{The Influence of Additional Parameters.} 
	To eliminate the influence of introducing more parameters,
	we replace every $1 \times 1$ convolution layer  
	in the model of the 1st row 
	with a sequence of two $3 \times 3$ convolution layers 
	($M$ sequences, $2 \times M$ convolution layers in total), 
	as shown in the 2nd row of Table.~\ref{tab:cii}.
	With more convolutional layers, 
	the overall performances are slightly improved (the 2nd \emph{v.s.} 1st rows).
	However, there is still a large margin compared to CII 
	(the 2nd \emph{v.s.} 3rd rows).
	This phenomenon demonstrates that more parameters do not 
	necessarily mean better performance. 
	A similar conclusion can be drawn by comparing 
	the 4th with the 3rd rows in Table.~\ref{tab:cii}.
	\item
	\textbf{Visualization of the Features.} 
	The utilization of CII enables the
	previous independent multi-scale 
	features to interact with each other before being 
	used to build the top-down pathway. 
	We visualize the intermediate features after the bottom-up pathway 
	and before the top-down pathway in Fig.~\ref{fig:vis_cii}. 
	Columns (a,b) are obtained from the model 
	described in the 2nd row of Table.~\ref{tab:cii},
	while columns (c,d) are from the 3rd row.
	As can be seen, with the help of CII, 
	the lower-level features tend to highlight 
	the whole building while the indistinct higher-level features 
	become more confident on the building's boundaries.
	On the opposite, without CII, the feature maps are visually identical.
\end{itemize}

\begin{table}[t]
	\centering
	\small
	\renewcommand{\arraystretch}{1.1}
	\setlength\tabcolsep{0.57mm}
	\begin{tabular}{c|c|c|ccc|ccc} \whline{1pt}
	  Kernel & \multirow{2}*{No.} & \multirow{2}*{Share} & \multicolumn{3}{c|}{DUT-OMRON \cite{yang2013saliency}} 
	  & \multicolumn{3}{c}{DUTS-TE \cite{wang2017learning}} \\
	  \cline{4-6} \cline{7-9} 
	  Size & & & $F_\beta$~$\uparrow$ & MAE~$\downarrow$ & $S_\alpha$~$\uparrow$ & $F_\beta$~$\uparrow$ & MAE~$\downarrow$ & $S_\alpha$~$\uparrow$  \\ \whline{0.5pt} 
	  $1 \times 1$ & $1 \times M$ & \xmark & 0.801 & 0.075 & 0.816 & 0.848 & 0.060 & 0.851 \\
	  $3 \times 3$ & $2 \times M$ & \xmark & 0.804 & 0.075 & 0.818 & 0.849 & 0.059 & 0.852 \\
	  $3 \times 3$ & $2 \times 1$ & \cmark & 0.810 & 0.069 & 0.822 & \sorb{0.869} & 0.050 & \sorb{0.870} \\
	  $3 \times 3$ & $4 \times 1$ & \cmark & \sorb{0.812} & \sorb{0.066} & \sorb{0.825} & 0.868 & \sorb{0.049} & \sorb{0.870} \\
	  \whline{1pt}
	\end{tabular} 
	\vspace{1pt} 
	\caption{Ablation analysis of the proposed CII strategy on two popular datasets. 
	$M$ refers to the number of connections between the bottom-up and top-down pathways.
	The best result in each column is highlighted in \sorb{red}.}
	\label{tab:cii}
	\vspace{-4pt}
  \end{table}

\myPara{Effectiveness of RGC: }
%
%
To prove the effectiveness of RGC, we conduct a series 
of ablation experiments comparing different settings of  
the centralized information interactors ($InI_i$ in Equ:~\ref{equ:cii}).
Note that in the following experiments, except for the 
information interactor itself, all other configurations 
are identical to the 4th row in Table.~\ref{tab:cii}. 
%

%
\begin{itemize}
	\item
	\textbf{RGC \emph{v.s.} $3 \times 3$ Convs.} 
	As can be seen from the 2nd row compared 
	to the 1st row in \tabref{tab:rgc}, 
	with the help of the RGC module, 
	better overall performances are reached.
	This proves the effectiveness of introducing an extra branch to 
	calibrate the local feature with 
	its relative global information within each specific input scale.
	\item
	\textbf{RGC \emph{v.s.} PPM.}  
	To investigate the effectiveness of previous multi-scale modules,
	we migrate the successful PPM \cite{zhao2016pyramid} module 
	into CII by directly replacing the RGC module with it.
	However, as shown in the 4th row in Table.~\ref{tab:rgc}, 
	PPM even performs slightly worse than those with 
	only $3 \times 3$ convolution layers (the 1st row).
	These numerical results indicate that the previous successful 
	multi-scale modules may not necessarily succeed in CII.
	\item
	\textbf{$\text{RGC}^\text{\dag}$ \emph{v.s.} RGC.} 
	$\text{RGC}^\text{\dag}$ changes the input of its 
	right (relative global) branch to be 
	the feature of the succeeding stage ($B_i$ to $B_{i+1}$). 
	By comparing the 3rd to 2nd rows in Table.~\ref{tab:rgc}, 
	we can see that the performances are further promoted.
	We also try $\text{PPM}^\text{\dag}$ (the 5th row), which is obtained 
	by changing the inputs of the branches with spatial interpolation operations
	in PPM to be the feature of the succeeding stage.
	However, the results are not satisfactory.
	%
	%
	These phenomenons show a promising direction of developing 
	new multi-scale modules better cooperating with CII.
	%
\end{itemize}

\begin{table}[t]
	\centering
	\small
	\renewcommand{\arraystretch}{1.1}
	\setlength\tabcolsep{1.2mm}
	\begin{tabular}{c|ccc|ccc} \whline{1pt}
	  Information & \multicolumn{3}{c|}{DUT-OMRON \cite{yang2013saliency}} & \multicolumn{3}{c}{DUTS-TE \cite{wang2017learning}} \\
	  \cline{2-4} \cline{5-7} 
	  Interactor & $F_\beta$~$\uparrow$ & MAE~$\downarrow$ & $S_\alpha$~$\uparrow$ & $F_\beta$~$\uparrow$ & MAE~$\downarrow$ & $S_\alpha$~$\uparrow$  \\ \whline{0.7pt}
	  $3 \times 3$ Convs & 0.812 & 0.066 & 0.825 & 0.868 & 0.049 & 0.870 \\ \whline{0.5pt}
	  RGC~ & 0.820 & 0.061 & 0.826 & 0.873 & 0.045 & 0.870 \\
	  $\text{RGC}^\text{\dag}$ & \sorb{0.824} & \sorb{0.058} & \sorb{0.828} & \sorb{0.878} & \sorb{0.042} & \sorb{0.874} \\ \whline{0.5pt}
	  PPM \cite{zhao2016pyramid} & 0.810 & 0.066 & 0.826 & 0.868 & 0.047 & 0.868 \\
	  $\text{PPM}^\text{\dag}$ & 0.803 & 0.070 & 0.821 & 0.866 & 0.047 & 0.869 \\
	  \whline{1pt}
	\end{tabular} 
	\vspace{1pt} 
	\caption{Ablation analysis of the proposed RGC module on two popular datasets.
	The 1st row uses a sequence of four $3 \times 3$ convolution layers 
	as the information interactor 
	(the same as the 4th row in Table.~\ref{tab:cii}). The best result in
	each column is highlighted in \sorb{red}.}
	\label{tab:rgc}
	\vspace{-4pt}
  \end{table}

\subsection{Comparisons to the State-of-the-Arts}
In this section, 
we compare our proposed approach with 15  
previous state-of-the-art methods, 
including DCL~\cite{li2016deep}, 
DSS~\cite{hou2016deeply},
Amulet~\cite{zhang2017amulet},
RAS~\cite{ras2020tip},
PiCANet~\cite{liu2018picanet},
AFNet~\cite{feng2019attentive}, 
MLMS~\cite{wu2019mutual}, 
JDFPR~\cite{xu2019deepcrf}, 
PAGE~\cite{wang2019salient},
CPD~\cite{wu2019cascaded},  BASNet~\cite{qin2019basnet}, 
PoolNet~\cite{liu2019simple}, CSNet~\cite{gao2020sod100k}, ITSD~\cite{zhou2020interactive},
and MINet~\cite{pang2020multi}.  
%
%
For fair comparisons,
the saliency maps of other methods are 
generated by the original codes released by the corresponding authors 
or directly provided by them.
We evaluate all the results with the same evaluation codes.

\begin{table*}[tp!] 
	\centering
    \small
    \renewcommand{\arraystretch}{1.15}
    \renewcommand{\tabcolsep}{0.35mm}
    \begin{tabular}{l|c|c|ccc|ccc|ccc|ccc|ccc}
        \whline{1pt}
        \multirow{2}*{Method}  & Params & FLOPs  &
        \multicolumn{3}{c|}{ECSSD \cite{yan2013hierarchical}} & \multicolumn{3}{c|}{PASCAL-S \cite{li2014secrets}} & \multicolumn{3}{c|}{DUT-OMRON \cite{yang2013saliency}} & \multicolumn{3}{c|}{HKU-IS \cite{li2015visual}} & \multicolumn{3}{c}{DUTS-TE \cite{wang2017learning}} 
        \\ \cline{4-18} 
        & (M) & (G) & $F_\beta$~$\uparrow$ & MAE~$\downarrow$ & $S_\alpha$~$\uparrow$ & $F_\beta$~$\uparrow$ & MAE~$\downarrow$ & $S_\alpha$~$\uparrow$ & $F_\beta$~$\uparrow$ & MAE~$\downarrow$ & $S_\alpha$~$\uparrow$ & $F_\beta$~$\uparrow$ & MAE~$\downarrow$ & $S_\alpha$~$\uparrow$ & $F_\beta$~$\uparrow$ & MAE~$\downarrow$ & $S_\alpha$~$\uparrow$ 
        \\ \hline
        $\text{DCL}_\text{16}$~\cite{li2016deep} & 66.25 & - & 0.898 & 0.078 & 0.873 & 0.805 & 0.115 & 0.800 & 0.733 & 0.094 & 0.762 & 0.893 & 0.063 & 0.871 & 0.786 & 0.081 & 0.803 \\
        $\text{DSS}_\text{17}$~\cite{hou2016deeply} & 62.23 & 52.20 & 0.908 & 0.062 & 0.884 & 0.821 & 0.101 & 0.804 & 0.760 & 0.074 & 0.789 & 0.900 & 0.050 & 0.881  & 0.813 & 0.065 & 0.826 \\
        $\text{Amulet}_\text{17}$~\cite{zhang2017amulet} & 33.16 & 20.70 & 0.913 & 0.060 & 0.881 & 0.826 & 0.092 & 0.816 & 0.737 & 0.083 & 0.784 & 0.889 & 0.052 & 0.866 & 0.773 & 0.075 & 0.800 \\
		$\text{RAS}_\text{18}$~\cite{ras2020tip}  & 20.23 & 21.24 & 0.921 & 0.056 & 0.893 & 0.829 & 0.101 & 0.799 & 0.786 & 0.062 & 0.814 & 0.913 & 0.045 & 0.887 & 0.831 & 0.059 & 0.839 \\ 
        $\text{PiCANet}_\text{18}$~\cite{liu2018picanet}  & 47.22 & 54.06 & 0.935 & 0.047 & 0.917 & 0.864 & 0.075 & 0.854 & 0.820 & 0.064 & 0.830 & 0.920 & 0.044 & 0.904 & 0.863 & 0.050 & 0.868 \\ 
        $\text{AFNet}_\text{19}$~\cite{feng2019attentive} & 25.78 & - & 0.936 & 0.042 & 0.914 & 0.861 & 0.070 & 0.849 & 0.820 & 0.057 & 0.825 & 0.926 & 0.036 & 0.906 & 0.867 & 0.045 & 0.867 \\ 
        $\text{MLMS}_\text{19}$~\cite{wu2019mutual}  & 74.38 & 58.18 & 0.930 & 0.045 & 0.911 & 0.853 & 0.074 & 0.844 & 0.793 & 0.063 & 0.809 & 0.922 & 0.039 & 0.907 & 0.854 & 0.048 & 0.862 \\ 
        $\text{JDFPR}_\text{19}$~\cite{xu2019deepcrf}  & 87.61 & 42.96 & 0.928 & 0.049 & 0.907 & 0.854 & 0.082 & 0.841 & 0.802 & 0.057 & 0.821 & - & - & -  & 0.833 & 0.058 & 0.836 \\ 
        $\text{PAGE}_\text{19}$~\cite{wang2019salient} & - & - & 0.931 & 0.042 & 0.912 & 0.848 & 0.076 & 0.842 & 0.791 & 0.062 & 0.825 & 0.920 & 0.036 & 0.904  & 0.838 & 0.051 & 0.855 \\ 
        \rowcolor{blue!9} $\text{CPD}_\text{19}$~\cite{wu2019cascaded}  & 47.85 & 7.23 & 0.939 & 0.037 & 0.918 & 0.859 & 0.071 & 0.848 & 0.796 & 0.056 & 0.825 & 0.925 & 0.034 & 0.907 & 0.865 & 0.043 & 0.869 \\ 
        $\text{BASNet}_\text{19}$~\cite{qin2019basnet}  & 87.06 & 97.65 & 0.942 & 0.037 & 0.916 & 0.857 & 0.076 & 0.838 & 0.811 & 0.057 & 0.836 & 0.930 & 0.033 & 0.908 & 0.860 & 0.047 & 0.866 \\ \whline{0.5pt}
        \rowcolor{blue!7} \textbf{Ours(ResNet-18)} & 11.89 & 6.49 & 0.941 & 0.039 & 0.916 & 0.868 & 0.068 & 0.851 & 0.824 & 0.058 & 0.828 & 0.933 & 0.032 & 0.912  & 0.878 & 0.042 & 0.874 \\ \whline{0.5pt}
        $\text{PoolNet}_\text{19}$~\cite{liu2019simple} & 68.26 & 38.19 & 0.944 & 0.039 & 0.921 & 0.865 & 0.075 & 0.850 & \sobb{0.830} & \sobb{0.055} & 0.836 & 0.934 & 0.032 & 0.917 & 0.886 & 0.040 & 0.883 \\
        $\text{CSNet}_\text{20}$~\cite{gao2020sod100k}  & 36.37 & 11.75 & 0.944 & 0.038 & 0.921 & 0.866 & 0.073 & 0.851 & 0.821 & \sobb{0.055} & 0.831 & 0.930 & 0.033 & 0.911 & 0.881 & 0.040 & 0.879 \\
        \rowcolor{gray!23} $\text{ITSD}_\text{20}$~\cite{zhou2020interactive}  & 26.47 & 9.67 & \sobb{0.947} & \sobb{0.035} & \sobb{0.925} & 0.871 & 0.066 & \sobb{0.859} & 0.823 & 0.061 & \sorb{0.840} & 0.933 & 0.031 & 0.916 & 0.883 & 0.041 & \sobb{0.885} \\
        $\text{MINet}_\text{20}$~\cite{pang2020multi}  & 162.38 & 42.73 & \sobb{0.947} & \sorb{0.034} & \sobb{0.925} & \sobb{0.874} & \sobb{0.064} & 0.856 & 0.826 & 0.056 & 0.833 & \sobb{0.936} & \sorb{0.028} & \sorb{0.920} & \sobb{0.888} & \sobb{0.037} & 0.884 \\
        \whline{0.5pt}
        
        \rowcolor{gray!20} \textbf{Ours(ResNet-50)} & 24.48 & 8.88  & \sorb{0.950} & \sorb{0.034} & \sorb{0.926} & \sorb{0.882} & \sorb{0.062} & \sorb{0.865} & \sorb{0.831} & \sorb{0.054} & \sobb{0.839} & \sorb{0.939} & \sobb{0.029} & \sobb{0.919} & \sorb{0.890} & \sorb{0.036} & \sorb{0.888} \\
		\whline{1pt}
	\end{tabular}
	\vspace{8pt}
	\caption{
		Quantitative comparisons on five widely used datasets.
		The best and second-best results in each column are highlighted in \sorb{red} and \sobb{blue}, respectively. We also show the results of the ResNet-18 version of our approach. 
		Lines in gray or blue mean methods of similar computational complexities (FLOPs), respectively. 
		The FLOPs of all approaches 
		are measured with an input image size of $224 \times 224$.
	}\label{tab:results}
\end{table*}

\renewcommand{\addFig}[1]{\includegraphics[width=0.318\linewidth]{prs/#1.pdf}}
\begin{figure*}[t]
  \centering
  \footnotesize
  \setlength\tabcolsep{1.3mm}
  \renewcommand\arraystretch{1.2}
  \begin{tabular}{ccc}
    \addFig{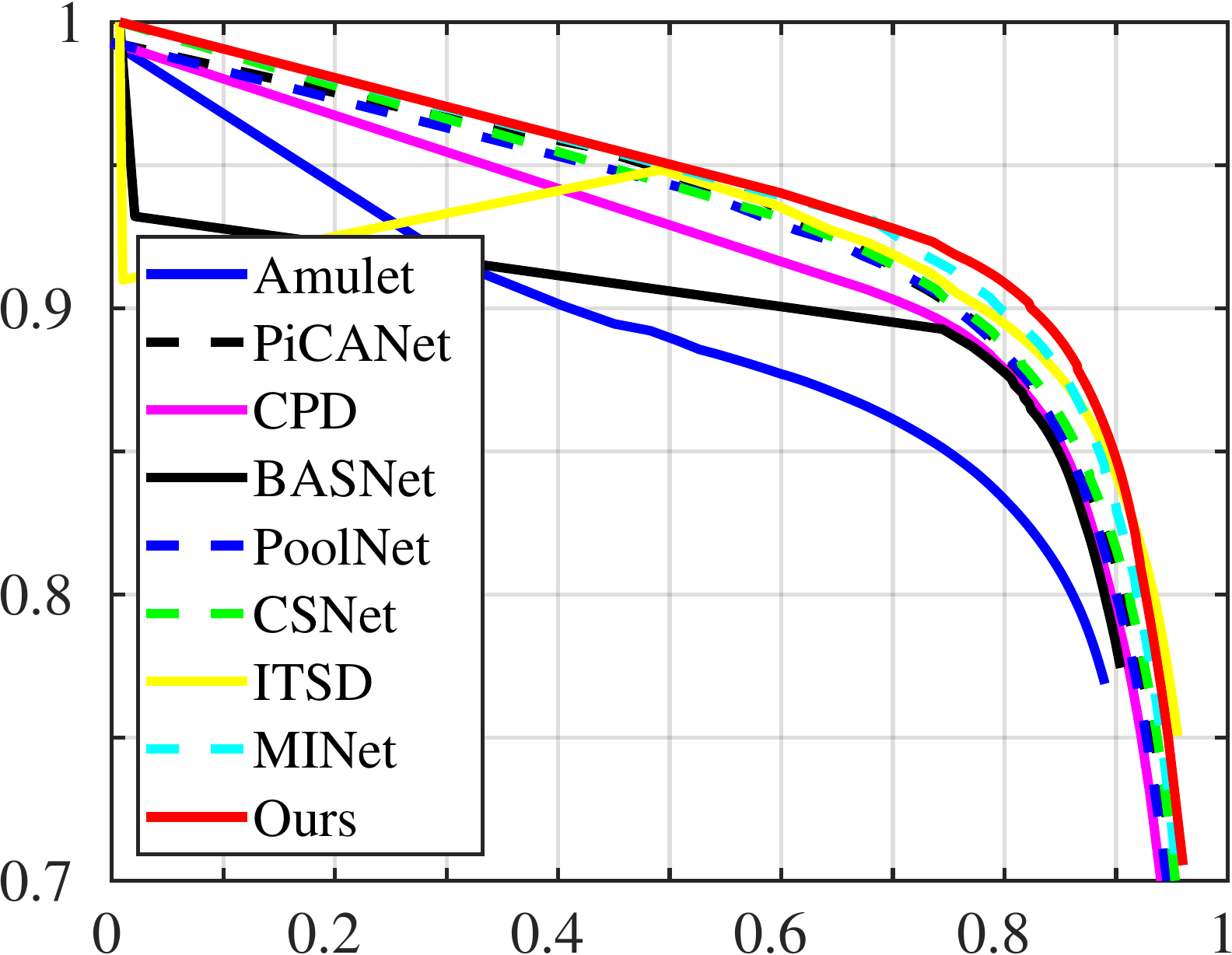} & \addFig{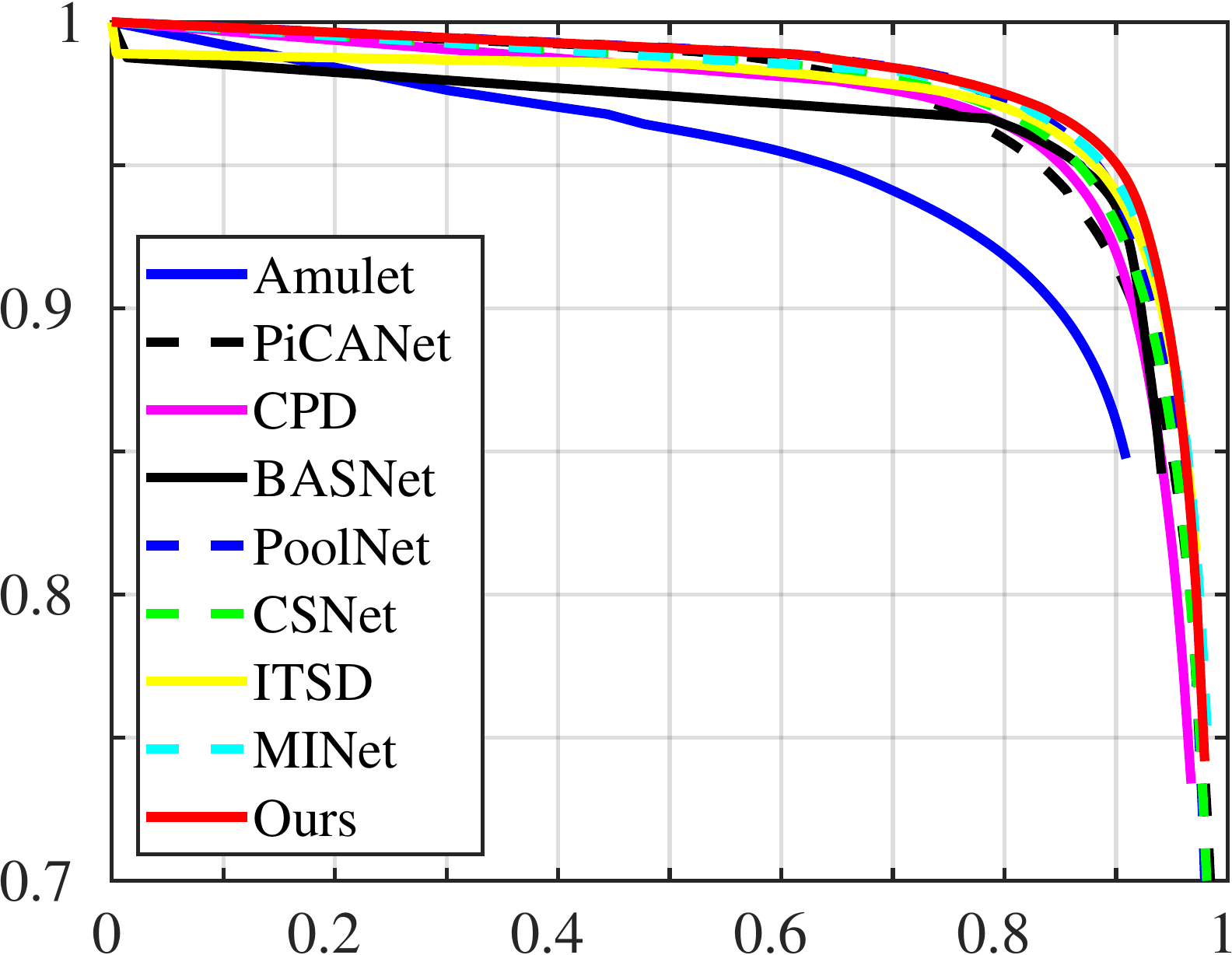} & \addFig{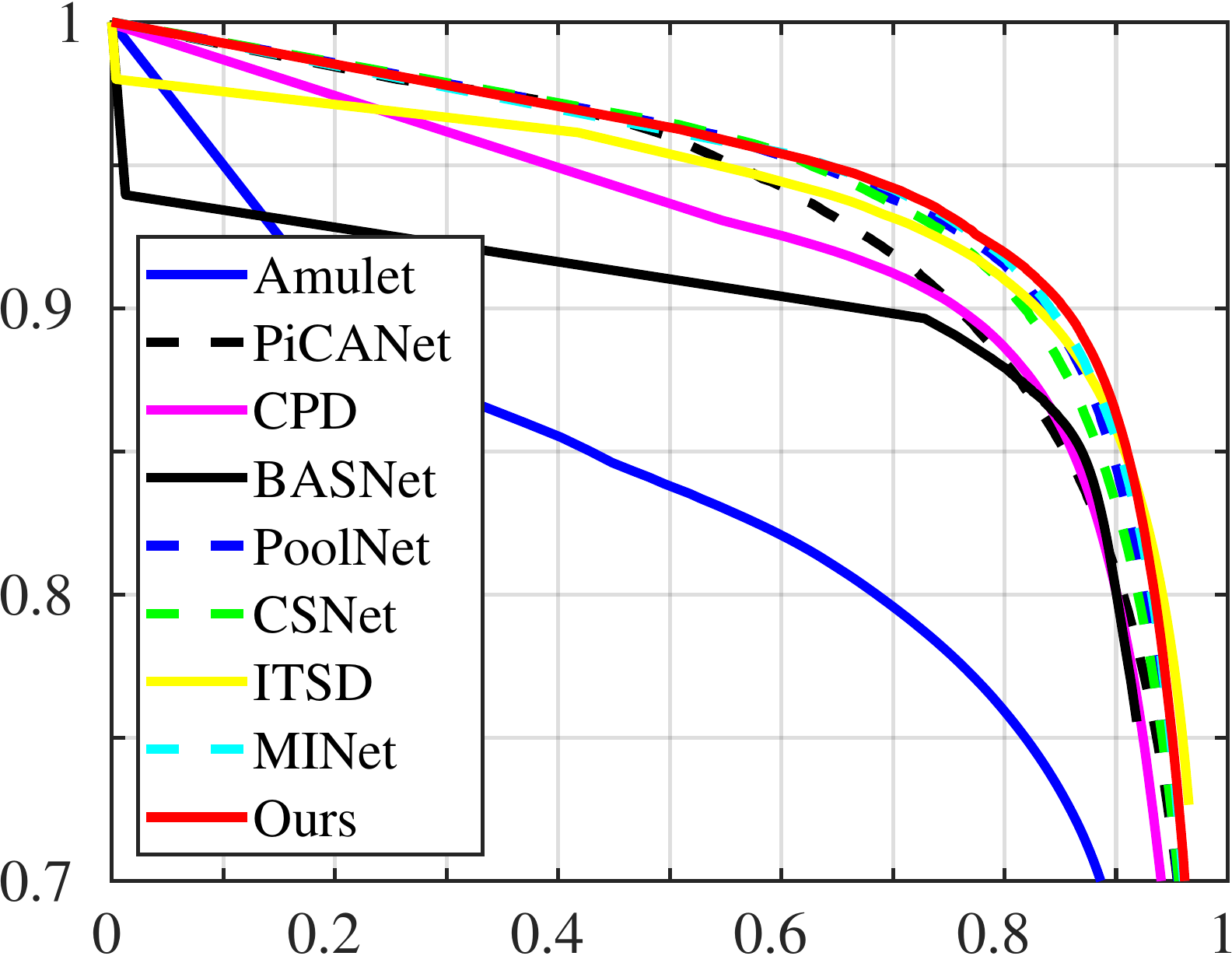} \\
    (a) PASCAL-S~\cite{li2014secrets} & (b) HKU-IS~\cite{li2015visual} & (c) DUTS-TE~\cite{wang2017learning} \\
  \end{tabular}
  \caption{Precision (vertical axis) recall (horizontal axis) curves on
    three popular salient object detection datasets. }
  \label{fig:prs}
\end{figure*}

\myPara{Quantitative Comparisons }
Quantitative results are listed in Table.~\ref{tab:results}.
As can be seen, the ResNet-50 version of our proposed approach 
achieves the best performances on most of the datasets and metrics. 
We only introduce 0.97M (4.1\%) additional parameters 
compared to the ResNet-50 backbone (23.51M).
Especially, the RGC module only occupies 0.22M (0.9\%) parameters 
while the other 0.75M (3.2\%)
parameters are essential components to construct the top-down pathway.
When compared to the previous most efficient method ITSD \cite{zhou2020interactive} 
(based on ResNet-50), 
our method achieves obvious better results on most of the datasets while  
requiring 67\% less additional parameters (0.97M \emph{v.s.} 2.96M) 
and 14\% less additional FLOPs (4.74G \emph{v.s.} 5.53G)

We also show the results of the ResNet-18 version of our approach in Table.~\ref{tab:results}.
Surprisingly, our ResNet-18 version still outperforms most of the 
previous methods based on the more powerful ResNet-50 network 
with even fewer parameters and FLOPs.
Besides the numerical comparison, 
we also show the PR curves
on three datasets in \figref{fig:prs}.
It is obvious that the PR curves of our approach (red ones) surpass 
almost all other methods under most thresholds.
This verifies the robustness of our approach.

\myPara{Visual Comparisons: }
In Fig.~\ref{fig:vis_comps}, we show some representative 
examples to evaluate our proposed method visually.
It can be easily observed that our proposed method can 
not only accurately highlight the salient objects but also 
segment them out integrally in almost 
all circumstances. 
Different from \cite{liu2020dynamic,wang2019salient,zhao2019egnet,wu2019stacked}, 
our approach requires no extra supervision on the edge 
areas. 
It demonstrates our method's effectiveness in 
augmenting the information interaction among the multi-scale features 
and generating semantically stronger and 
positionally more precise features.

\section{Conclusions and Future Work}
This paper advances the classic U-shape structure by 
centralizing the previous independent connections 
between its bottom-up and top-down 
pathways to encourage the information interaction among 
multi-scale features.
To show that this centralized information interaction (CII) 
strategy is feasible, we further propose a relative global calibration (RGC) 
module to cooperate with it.
By combining CII and RGC into the classic U-shape architecture, 
we show that our proposed approach can surpass 
all previous state-of-the-art methods on 
five widely-used salient object detection
benchmarks with a handful number of additional parameters and FLOPs.
Our proposed strategy and module 
are independent of the designs of the bottom-up and top-down 
pathways in the U-shape structures
and hence can be flexibly
applied to any U-shape based models.
Also, the proposed RGC shows a promising direction of developing 
new multi-scale modules better cooperating with the proposed CII strategy.
\clearpage
{\small
    \bibliographystyle{ieee_fullname}
    \bibliography{egbib}

\begin{thebibliography}{10}\itemsep=-1pt

\bibitem{BorjiCVM2019}
Ali Borji, Ming-Ming Cheng, Qibin Hou, Huaizu Jiang, and Jia Li.
\newblock Salient object detection: A survey.
\newblock {\em Computational Visual Media}, 5(2):117--150, 2019.

\bibitem{borji2015salient}
Ali Borji, Ming-Ming Cheng, Huaizu Jiang, and Jia Li.
\newblock Salient object detection: A benchmark.
\newblock {\em {IEEE TIP}}, 24(12):5706--5722, 2015.

\bibitem{chen2017deeplab}
Liang-Chieh Chen, George Papandreou, Iasonas Kokkinos, Kevin Murphy, and Alan~L
  Yuille.
\newblock Deeplab: Semantic image segmentation with deep convolutional nets,
  atrous convolution, and fully connected crfs.
\newblock {\em {IEEE TPAMI}}, 2017.

\bibitem{ras2020tip}
Shuhan Chen, Xiuli Tan, Ben Wang, Huchuan Lu, Xuelong Hu, and Yun Fu.
\newblock Reverse attention-based residual network for salient object
  detection.
\newblock {\em {IEEE TIP}}, 29:3763--3776, 2020.

\bibitem{cheng2015global}
Ming-Ming Cheng, Niloy~J. Mitra, Xiaolei Huang, Philip H.~S. Torr, and Shi-Min
  Hu.
\newblock Global contrast based salient region detection.
\newblock {\em {IEEE TPAMI}}, 37(3):569--582, 2015.

\bibitem{cheng2010repfinder}
Ming-Ming Cheng, Fang-Lue Zhang, Niloy~J Mitra, Xiaolei Huang, and Shi-Min Hu.
\newblock Repfinder: finding approximately repeated scene elements for image
  editing.
\newblock {\em {ACM TOG}}, 29(4):83, 2010.

\bibitem{craye2016environment}
Celine Craye, David Filliat, and Jean-Fran{\c{c}}ois Goudou.
\newblock Environment exploration for object-based visual saliency learning.
\newblock In {\em {ICRA}}, pages 2303--2309, 2016.

\bibitem{fan2017structure}
Deng-Ping Fan, Ming-Ming Cheng, Yun Liu, Tao Li, and Ali Borji.
\newblock {Structure-measure: A New Way to Evaluate Foreground Maps}.
\newblock In {\em {ICCV}}, pages 4548--4557, 2017.

\bibitem{feng2019attentive}
Mengyang Feng, Huchuan Lu, and Errui Ding.
\newblock Attentive feedback network for boundary-aware salient object
  detection.
\newblock In {\em {CVPR}}, 2019.

\bibitem{fu2019dual}
Jun Fu, Jing Liu, Haijie Tian, Yong Li, Yongjun Bao, Zhiwei Fang, and Hanqing
  Lu.
\newblock Dual attention network for scene segmentation.
\newblock In {\em {CVPR}}, pages 3146--3154, 2019.

\bibitem{gao2020sod100k}
Shang-Hua Gao, Yong-Qiang Tan, Ming-Ming Cheng, Chengze Lu, Yunpeng Chen, and
  Shuicheng Yan.
\newblock Highly efficient salient object detection with 100k parameters.
\newblock In {\em {ECCV}}, 2020.

\bibitem{ghiasi2019fpn}
Golnaz Ghiasi, Tsung-Yi Lin, and Quoc~V Le.
\newblock Nas-fpn: Learning scalable feature pyramid architecture for object
  detection.
\newblock In {\em {CVPR}}, pages 7036--7045, 2019.

\bibitem{He2016}
Kaiming He, Xiangyu Zhang, Shaoqing Ren, and Jian Sun.
\newblock Deep residual learning for image recognition.
\newblock In {\em {CVPR}}, 2016.

\bibitem{hong2015online}
Seunghoon Hong, Tackgeun You, Suha Kwak, and Bohyung Han.
\newblock Online tracking by learning discriminative saliency map with
  convolutional neural network.
\newblock In {\em ICML}, pages 597--606, 2015.

\bibitem{hou2016deeply}
Qibin Hou, Ming-Ming Cheng, Xiaowei Hu, Ali Borji, Zhuowen Tu, and Philip Torr.
\newblock Deeply supervised salient object detection with short connections.
\newblock {\em {IEEE TPAMI}}, 41(4):815--828, 2019.

\bibitem{hou2018self}
Qibin Hou, Peng-Tao Jiang, Yunchao Wei, and Ming-Ming Cheng.
\newblock Self-erasing network for integral object attention.
\newblock In {\em {NeurIPS}}, 2018.

\bibitem{huang2020ccnet}
Zilong Huang ; Xinggang Wang ; Yunchao Wei ; Lichao Huang ; Humphrey Shi ;
  Wenyu Liu ; Thomas~S. Huang.
\newblock Ccnet: Criss-cross attention for semantic segmentation.
\newblock {\em {IEEE TPAMI}}, pages 1--1, 2020.

\bibitem{ioffe2015batch}
Sergey Ioffe and Christian Szegedy.
\newblock Batch normalization: Accelerating deep network training by reducing
  internal covariate shift.
\newblock In {\em ICML}, 2015.

\bibitem{jiang2013salient}
Huaizu Jiang, Jingdong Wang, Zejian Yuan, Yang Wu, Nanning Zheng, and Shipeng
  Li.
\newblock Salient object detection: A discriminative regional feature
  integration approach.
\newblock In {\em {CVPR}}, pages 2083--2090, 2013.

\bibitem{krizhevsky2012imagenet}
Alex Krizhevsky, Ilya Sutskever, and Geoffrey~E Hinton.
\newblock Imagenet classification with deep convolutional neural networks.
\newblock In {\em {NeurIPS}}, 2012.

\bibitem{li2017instance}
Guanbin Li, Yuan Xie, Liang Lin, and Yizhou Yu.
\newblock Instance-level salient object segmentation.
\newblock In {\em {CVPR}}, 2017.

\bibitem{li2015visual}
Guanbin Li and Yizhou Yu.
\newblock Visual saliency based on multiscale deep features.
\newblock In {\em {CVPR}}, pages 5455--5463, 2015.

\bibitem{li2016deep}
Guanbin Li and Yizhou Yu.
\newblock Deep contrast learning for salient object detection.
\newblock In {\em {CVPR}}, 2016.

\bibitem{li2013saliency}
Xiaohui Li, Huchuan Lu, Lihe Zhang, Xiang Ruan, and Ming-Hsuan Yang.
\newblock Saliency detection via dense and sparse reconstruction.
\newblock In {\em {ICCV}}, pages 2976--2983, 2013.

\bibitem{li2014secrets}
Yin Li, Xiaodi Hou, Christof Koch, James~M Rehg, and Alan~L Yuille.
\newblock The secrets of salient object segmentation.
\newblock In {\em {CVPR}}, pages 280--287, 2014.

\bibitem{li2020cross}
Zun Li, Congyan Lang, Junhao Liew, Qibin Hou, Yidong Li, and Jiashi Feng.
\newblock Cross-layer feature pyramid network for salient object detection.
\newblock {\em arXiv preprint arXiv:2002.10864}, 2020.

\bibitem{lin2017feature}
Tsung-Yi Lin, Piotr Doll{\'a}r, Ross~B Girshick, Kaiming He, Bharath Hariharan,
  and Serge~J Belongie.
\newblock Feature pyramid networks for object detection.
\newblock In {\em {CVPR}}, 2017.

\bibitem{liu2019auto}
Chenxi Liu, Liang-Chieh Chen, Florian Schroff, Hartwig Adam, Wei Hua, Alan~L
  Yuille, and Li Fei-Fei.
\newblock Auto-deeplab: Hierarchical neural architecture search for semantic
  image segmentation.
\newblock In {\em {CVPR}}, pages 82--92, 2019.

\bibitem{liu2020dynamic}
Jiang-Jiang Liu, Qibin Hou, and Ming-Ming Cheng.
\newblock Dynamic feature integration for simultaneous detection of salient
  object, edge and skeleton.
\newblock {\em {IEEE TIP}}, 29:8652--8667, 2020.

\bibitem{liu2019simple}
Jiang-Jiang Liu, Qibin Hou, Ming-Ming Cheng, Jiashi Feng, and Jianmin Jiang.
\newblock A simple pooling-based design for real-time salient object detection.
\newblock In {\em {CVPR}}, 2019.

\bibitem{liu2020improving}
Jiang-Jiang Liu, Qibin Hou, Ming-Ming Cheng, Changhu Wang, and Jiashi Feng.
\newblock Improving convolutional networks with self-calibrated convolutions.
\newblock In {\em {CVPR}}, 2020.

\bibitem{liu2016dhsnet}
Nian Liu and Junwei Han.
\newblock Dhsnet: Deep hierarchical saliency network for salient object
  detection.
\newblock In {\em {CVPR}}, 2016.

\bibitem{liu2018picanet}
Nian Liu, Junwei Han, and Ming-Hsuan Yang.
\newblock Picanet: Learning pixel-wise contextual attention for saliency
  detection.
\newblock In {\em {CVPR}}, pages 3089--3098, 2018.

\bibitem{liu2019learning}
Songtao Liu, Di Huang, and Yunhong Wang.
\newblock Learning spatial fusion for single-shot object detection.
\newblock {\em arXiv preprint arXiv:1911.09516}, 2019.

\bibitem{liu2018path}
Shu Liu, Lu Qi, Haifang Qin, Jianping Shi, and Jiaya Jia.
\newblock Path aggregation network for instance segmentation.
\newblock In {\em {CVPR}}, pages 8759--8768, 2018.

\bibitem{long2015fully}
Jonathan Long, Evan Shelhamer, and Trevor Darrell.
\newblock Fully convolutional networks for semantic segmentation.
\newblock In {\em {CVPR}}, pages 3431--3440, 2015.

\bibitem{luo2017non}
Zhiming Luo, Akshaya~Kumar Mishra, Andrew Achkar, Justin~A Eichel, Shaozi Li,
  and Pierre-Marc Jodoin.
\newblock Non-local deep features for salient object detection.
\newblock In {\em {CVPR}}, 2017.

\bibitem{nair2010rectified}
Vinod Nair and Geoffrey~E Hinton.
\newblock Rectified linear units improve restricted boltzmann machines.
\newblock In {\em ICML}, 2010.

\bibitem{pang2019libra}
Jiangmiao Pang, Kai Chen, Jianping Shi, Huajun Feng, Wanli Ouyang, and Dahua
  Lin.
\newblock Libra r-cnn: Towards balanced learning for object detection.
\newblock In {\em {CVPR}}, pages 821--830, 2019.

\bibitem{pang2020multi}
Youwei Pang, Xiaoqi Zhao, Lihe Zhang, and Huchuan Lu.
\newblock Multi-scale interactive network for salient object detection.
\newblock In {\em {CVPR}}, pages 9413--9422, 2020.

\bibitem{perazzi2012saliency}
Federico Perazzi, Philipp Kr{\"a}henb{\"u}hl, Yael Pritch, and Alexander
  Hornung.
\newblock Saliency filters: Contrast based filtering for salient region
  detection.
\newblock In {\em {CVPR}}, pages 733--740, 2012.

\bibitem{qiao2020detectors}
Siyuan Qiao, Liang-Chieh Chen, and Alan Yuille.
\newblock Detectors: Detecting objects with recursive feature pyramid and
  switchable atrous convolution.
\newblock {\em arXiv preprint arXiv:2006.02334}, 2020.

\bibitem{qin2019basnet}
Xuebin Qin, Zichen Zhang, Chenyang Huang, Chao Gao, Masood Dehghan, and Martin
  Jagersand.
\newblock Basnet: Boundary-aware salient object detection.
\newblock In {\em {CVPR}}, 2019.

\bibitem{ronneberger2015u}
Olaf Ronneberger, Philipp Fischer, and Thomas Brox.
\newblock U-net: Convolutional networks for biomedical image segmentation.
\newblock In {\em MICCAI}, pages 234--241, 2015.

\bibitem{su2019selectivity}
Jinming Su, Jia Li, Yu Zhang, Changqun Xia, and Yonghong Tian.
\newblock Selectivity or invariance: Boundary-aware salient object detection.
\newblock In {\em {ICCV}}, pages 3799--3808, 2019.

\bibitem{tan2020efficientdet}
Mingxing Tan, Ruoming Pang, and Quoc~V Le.
\newblock Efficientdet: Scalable and efficient object detection.
\newblock In {\em {CVPR}}, pages 10781--10790, 2020.

\bibitem{wang2017learning}
Lijun Wang, Huchuan Lu, Yifan Wang, Mengyang Feng, Dong Wang, Baocai Yin, and
  Xiang Ruan.
\newblock Learning to detect salient objects with image-level supervision.
\newblock In {\em {CVPR}}, pages 136--145, 2017.

\bibitem{wang2017stagewise}
Tiantian Wang, Ali Borji, Lihe Zhang, Pingping Zhang, and Huchuan Lu.
\newblock A stagewise refinement model for detecting salient objects in images.
\newblock In {\em {ICCV}}, pages 4019--4028, 2017.

\bibitem{wang2018detect}
Tiantian Wang, Lihe Zhang, Shuo Wang, Huchuan Lu, Gang Yang, Xiang Ruan, and
  Ali Borji.
\newblock Detect globally, refine locally: A novel approach to saliency
  detection.
\newblock In {\em {CVPR}}, pages 3127--3135, 2018.

\bibitem{wang2019survey}
Wenguan Wang, Qiuxia Lai, Huazhu Fu, Jianbing Shen, and Haibin Ling.
\newblock Salient object detection in the deep learning era: An in-depth
  survey.
\newblock {\em arXiv preprint arXiv:1904.09146}, 2019.

\bibitem{wang2019iterative}
Wenguan Wang, Jianbing Shen, Ming-Ming Cheng, and Ling Shao.
\newblock An iterative and cooperative top-down and bottom-up inference network
  for salient object detection.
\newblock In {\em {CVPR}}, 2019.

\bibitem{wang2019revisiting}
Wenguan Wang, Jianbing Shen, Jianwen Xie, Ming-Ming Cheng, Haibin Ling, and Ali
  Borji.
\newblock Revisiting video saliency prediction in the deep learning era.
\newblock {\em {IEEE TPAMI}}, 2019.

\bibitem{wang2019salient}
Wenguan Wang, Shuyang Zhao, Jianbing Shen, Steven C.~H. Hoi, and Ali Borji.
\newblock Salient object detection with pyramid attention and salient edges.
\newblock In {\em {CVPR}}, 2019.

\bibitem{F3Net}
Jun Wei, Shuhui Wang, and Qingming Huang.
\newblock F3net: Fusion, feedback and focus for salient object detection.
\newblock In {\em {AAAI}}, 2020.

\bibitem{wei2020label}
Jun Wei, Shuhui Wang, Zhe Wu, Chi Su, Qingming Huang, and Qi Tian.
\newblock Label decoupling framework for salient object detection.
\newblock In {\em {CVPR}}, pages 13025--13034, 2020.

\bibitem{wei2016stc}
Yunchao Wei, Xiaodan Liang, Yunpeng Chen, Xiaohui Shen, Ming-Ming Cheng, Jiashi
  Feng, Yao Zhao, and Shuicheng Yan.
\newblock Stc: A simple to complex framework for weakly-supervised semantic
  segmentation.
\newblock {\em {IEEE TPAMI}}, 2016.

\bibitem{woo2018cbam}
Sanghyun Woo, Jongchan Park, Joon-Young Lee, and In So~Kweon.
\newblock Cbam: Convolutional block attention module.
\newblock In {\em {ECCV}}, pages 3--19, 2018.

\bibitem{wu2019mutual}
Runmin Wu, Mengyang Feng, Wenlong Guan, Dong Wang, Huchuan Lu, and Errui Ding.
\newblock A mutual learning method for salient object detection with
  intertwined multi-supervision.
\newblock In {\em {CVPR}}, 2019.

\bibitem{wu2019cascaded}
Zhe Wu, Li Su, and Qingming Huang.
\newblock Cascaded partial decoder for fast and accurate salient object
  detection.
\newblock In {\em {CVPR}}, 2019.

\bibitem{wu2019stacked}
Zhe Wu, Li Su, and Qingming Huang.
\newblock Stacked cross refinement network for edge-aware salient object
  detection.
\newblock In {\em {ICCV}}, 2019.

\bibitem{xu2019auto}
Hang Xu, Lewei Yao, Wei Zhang, Xiaodan Liang, and Zhenguo Li.
\newblock Auto-fpn: Automatic network architecture adaptation for object
  detection beyond classification.
\newblock In {\em {ICCV}}, pages 6649--6658, 2019.

\bibitem{xu2019deepcrf}
Yingyue Xu, Dan Xu, Xiaopeng Hong, Wanli Ouyang, Rongrong Ji, Min Xu, and
  Guoying Zhao.
\newblock Structured modeling of joint deep feature and prediction refinement
  for salient object detection.
\newblock In {\em {ICCV}}, 2019.

\bibitem{yan2013hierarchical}
Qiong Yan, Li Xu, Jianping Shi, and Jiaya Jia.
\newblock Hierarchical saliency detection.
\newblock In {\em {CVPR}}, pages 1155--1162, 2013.

\bibitem{yang2013saliency}
Chuan Yang, Lihe Zhang, Huchuan Lu, Xiang Ruan, and Ming-Hsuan Yang.
\newblock Saliency detection via graph-based manifold ranking.
\newblock In {\em {CVPR}}, pages 3166--3173, 2013.

\bibitem{yang2018denseaspp}
Maoke Yang, Kun Yu, Chi Zhang, Zhiwei Li, and Kuiyuan Yang.
\newblock Denseaspp for semantic segmentation in street scenes.
\newblock In {\em {CVPR}}, pages 3684--3692, 2018.

\bibitem{yuan2018ocnet}
Yuhui Yuan and Jingdong Wang.
\newblock Ocnet: Object context network for scene parsing.
\newblock {\em arXiv preprint arXiv:1809.00916}, 2018.

\bibitem{zeng2019towards}
Yi Zeng, Pingping Zhang, Jianming Zhang, Zhe Lin, and Huchuan Lu.
\newblock Towards high-resolution salient object detection.
\newblock In {\em {ICCV}}, pages 7234--7243, 2019.

\bibitem{zhang2018bi}
Lu Zhang, Ju Dai, Huchuan Lu, You He, and Gang Wang.
\newblock A bi-directional message passing model for salient object detection.
\newblock In {\em {CVPR}}, pages 1741--1750, 2018.

\bibitem{zhang2017amulet}
Pingping Zhang, Dong Wang, Huchuan Lu, Hongyu Wang, and Xiang Ruan.
\newblock Amulet: Aggregating multi-level convolutional features for salient
  object detection.
\newblock In {\em {ICCV}}, 2017.

\bibitem{zhang2018progressive}
Xiaoning Zhang, Tiantian Wang, Jinqing Qi, Huchuan Lu, and Gang Wang.
\newblock Progressive attention guided recurrent network for salient object
  detection.
\newblock In {\em {CVPR}}, pages 714--722, 2018.

\bibitem{zhao2016pyramid}
Hengshuang Zhao, Jianping Shi, Xiaojuan Qi, Xiaogang Wang, and Jiaya Jia.
\newblock Pyramid scene parsing network.
\newblock In {\em {CVPR}}, 2017.

\bibitem{zhao2019egnet}
Jia-Xing Zhao, Jiang-Jiang Liu, Deng-Ping Fan, Yang Cao, Jufeng Yang, and
  Ming-Ming Cheng.
\newblock Egnet:edge guidance network for salient object detection.
\newblock In {\em {ICCV}}, 2019.

\bibitem{zhao2019pyramid}
Ting Zhao and Xiangqian Wu.
\newblock Pyramid feature attention network for saliency detection.
\newblock In {\em {CVPR}}, 2019.

\bibitem{GateNet2020}
Xiaoqi Zhao, Youwei Pang, Lihe Zhang, Huchuan Lu, and Lei Zhang.
\newblock Suppress and balance: A simple gated network for salient object
  detection.
\newblock In {\em {ECCV}}, 2020.

\bibitem{zhou2020interactive}
Huajun Zhou, Xiaohua Xie, Jian-Huang Lai, Zixuan Chen, and Lingxiao Yang.
\newblock Interactive two-stream decoder for accurate and fast saliency
  detection.
\newblock In {\em {CVPR}}, pages 9141--9150, 2020.

\bibitem{zoph2016neural}
Barret Zoph and Quoc~V Le.
\newblock Neural architecture search with reinforcement learning.
\newblock In {\em ICLR}, 2017.

\end{thebibliography}
}

\end{document}